\documentclass[11pt]{article}

\usepackage[final]{acl}

\usepackage{times}
\usepackage{latexsym}

\usepackage[T1]{fontenc}
\usepackage[utf8]{inputenc}
\usepackage{microtype, graphicx}

\title{JU\_NLP  at HinglishEval: Quality Evaluation of the Low-Resource Code-Mixed Hinglish Text}

\author{Prantik Guha$^{1}$ \and Rudra Dhar$^{2}$ \and Dipankar Das$^{3}$ \\
        Department of Computer Science \& Engineering, Jadavpur University, Kolkata, India\\
        \{$^{1}$prantikguha706, $^{2}$rudradharrd, $^{3}$dipankar.dipnil2005\}@gmail.com}

\begin{document}
\maketitle
\begin{abstract}
In this paper we describe a system submitted to the INLG 2022 Generation Challenge (GenChal) on Quality Evaluation of the Low-Resource Synthetically Generated Code-Mixed Hinglish Text. We implement a Bi-LSTM-based neural network model to predict the Average rating score and Disagreement score of the synthetic Hinglish dataset. In our models, we used word embeddings for English and Hindi data, and one hot encodings for Hinglish data. We achieved a F1 score of 0.11, and mean squared error of 6.0 in the average rating score prediction task. In the task of Disagreement score prediction, we achieve a F1 score of 0.18, and mean squared error of 5.0.
\end{abstract}

\section{Introduction}

In India, social media's enduring popularity has resulted in massive amounts of user-generated textual content. During a conversation, multilingual speakers frequently flip between languages. Speakers frequently talk in multiple languages, and often transliterate. Listeners may not always be able to keep up with the multilingual speakers. That's why we need automated systems for transliterated translations.

But we don't have a significant amount of transliterated translation data to train our models. So we might use synthetic data for this purpose. Synthetic data has become a common resource for a variety of applications. It may be required because of data unavailability, cost savings, security, or privacy concerns. Because synthetic data matches the statistical properties of production data, it can be used to train models, validate models, and evaluate performance. Machine learning models have now made it possible to create incredibly fast natural language generating systems by building and training a model.

Now the next challenge is to evaluate the data which is synthetically generated. In this paper we have introduced an algorithm to check the quality of the generated data. We have proposed a supervised learning model using multiple Bi-LSTM and dense layers to predict two types of scores (Average Rating score and Disagreement score). In this paper we are using the data from \citet{srivastava-singh-2021-hinge}.

This is a transliterated translation verification problem which essentially boils down to a task of document similarity evaluation. Document similarity evaluation is a well researched task in NLP. As \citet{article} suggests, various Machine learning techniques, and Natural Language Processing tools can be used for this purpose. \citet{unknown} shows us how hybrid models of LSTM's can be used for document similarity prediction. Some work has also been done in the multilingual senario, as in \citet{jrfm11010008}. However not much work has been done in transliterated translation verification, and certainly none has been done in the Indian domain. \citet{srivastava2020phinc} explains the challenges in both generating transliterated translations and evaluating it.

\section{Dataset}

\begin{figure}[h]
    \includegraphics[width=8cm, height=9cm]{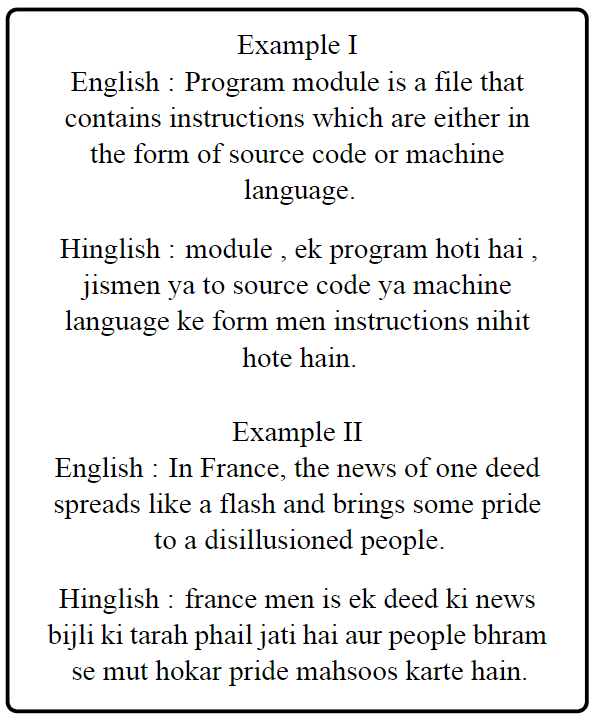}
    \caption{\label{figure:dataset-fig}
    Example from \cite{srivastava-singh-2021-hinge} data }
\end{figure}

The phenomena of code-mixing are the mingling of words and phrases from various languages in a single text or spoken utterance. Examples of code-mixed Hinglish sentences created from parallel Hindi and English utterances are shown in Fig-\ref{figure:dataset-fig}.

In this shared task, there are two subtasks for evaluating the quality of the code-mixed Hinglish text in this common task \cite{srivastava-singh-2021-quality}. In the first sub-task, they proposed using a scale of 1–10 to determine the quality of Hinglish content. They want to figure out what elements influence text quality, so high-quality code-mixed text generating systems can be created. The second sub-task is to predict how much the two annotators who annotated the synthetically generated Hinglish sentences differ on a scale of 0–9. Various factors influence human disagreement.

The dataset consists of five columns (English, Hindi, Hinglish, Average Rating, Disagreement). Highlish sentences are generated using two rule-based algorithms (i.e., WAC and PAC). For the two rating columns (Average Rating \& Disagreement) each sentence is rated on a scale of 1(low-quality) to 10 (high-quality) by two annotators.  The quality of the synthetically generated sentences is calculated by rounding off the average of the two human ratings and using this score (in the range of 1-10) in the Average rating column. And the Disagreement score is calculated by the absolute difference of the two human ratings as the disagreement score (in the range of 0-9).

\section{System Description}

\begin{figure*}[t]
    \includegraphics[width=\textwidth]{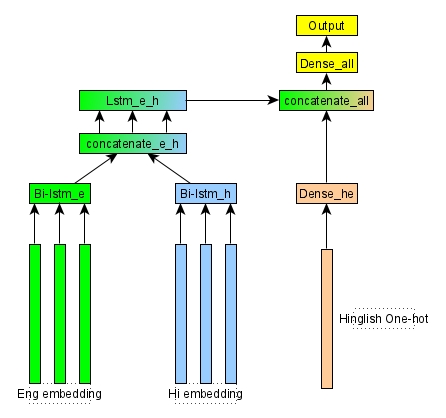}
    \caption{\label{figure:system}
    system architecture}
\end{figure*}

We used a sequence of Glove embeddings as input for English and Hindi sentences. However, for Hinglish sentences we used one hot vector as inputs. We fed the English and Hindi embeddings to separate Bi-lstm's[l-e, l-h], and retrieved sequence output from them. To capture the word sequences of different Hindi and English sentences we have used two different LSTMs. Then we concatenated these 2 outputs and passed it through another Lstm layer to get a fixed (not sequence) vector output [l-h-e]. 

We fed the one hot vector from the Hinglish data to a dense layer and received a vector output [d-he]. Since one hot vector does not capture the sequential information, we have used a dense layer. We then concatenated these two [l-h-e and d-he] vectors, and passed it through a dense layer to get a final class (score between 1 to 10).  We used the same model for both the tasks. Please refer to Fig-\ref{figure:system} for complete system architecture.

\section{Training}

\begin{table}[t]
\centering
\begin{tabular}{p{0.15\linewidth}|p{0.2\linewidth}|p{0.2\linewidth}|p{0.25\linewidth}}
\hline
\textbf{No. of data} & \textbf{F1-Score} & \textbf{Cohen's Kappa} & \textbf{Mean Squared Error}\\
\hline
395 & 0.09899 & -0.01521 & 6.00\\ 
\hline
\end{tabular}
\caption{\label{Sub-Task-1-Validation}
This result is obtained from 395 validation data for Sub-task 1(Average rating score)
}
\end{table}

\begin{table}[t]
\centering
\begin{tabular}{p{0.2\linewidth}|l|p{0.4\linewidth}}
\hline
\textbf{No. of data} & \textbf{F1-Score} & \textbf{Mean Squared Error}\\
\hline
395 & 0.21622 & 5.00\\ 
\hline
\end{tabular}
\caption{\label{Sub-Task-2-Validation}
This result is obtained from 395 validation data for Sub-task 2(Disagreement score)
}
\end{table}

On a total of 2766 training data points, we train the LSTM model using the Adam optimizer with a batch size of 32. Started with loss of 0.1810 \& accuracy of 0.9658. In the final epoch loss was 0.0300 \& accuracy was 0.9864.

In this phase, we validated the input using our developed model. For this phase the total available data was 395. We have validated our model for both Average Rating as well as Disagreement.On 395 data we validated our system to predict Average rating for corresponding inputs. Please refer to Table: \ref{Sub-Task-1-Validation} for detailed results related to this validation. On 395 data we validated our system to predict Disagreement score for corresponding inputs. Please refer to Table: \ref{Sub-Task-2-Validation} for detailed results related to this validation.

\section{Test}

\begin{table}[t]
\centering
\begin{tabular}{{p{0.15\linewidth}|p{0.2\linewidth}|p{0.2\linewidth}|p{0.25\linewidth}}}
\hline
\textbf{No. of data} & \textbf{F1-Score} & \textbf{Cohen's Kappa} & \textbf{Mean Squared Error}\\
\hline
791 & 0.11582 & 0.00337 & 6.00\\ 
\hline
\end{tabular}
\caption{\label{Sub-Task-1-Test}
This result is obtained from 791 test data for Sub-task 1(Average rating score)
}
\end{table}

\begin{table}[t]
\centering
\begin{tabular}{p{0.2\linewidth}|l|p{0.4\linewidth}}
\hline
\textbf{No. of data} & \textbf{F1-Score} & \textbf{Mean Squared Error}\\
\hline
791 & 0.18331 & 5.00\\ 
\hline
\end{tabular}
\caption{\label{Sub-Task-2-Test}
This result is obtained from 791 test data for Sub-task 2(Disagreement score)
}
\end{table}

In this phase, our developed model gets tested on test data. For this phase the total available data was 791. Model was tested for both Average Rating as well as Disagreement.

On 791 data, our system is able to predict Average rating for corresponding inputs. Please refer to Table: \ref{Sub-Task-1-Test} for detailed results related to this validation. On 791 data, our system is able to predict Disagreement score for corresponding inputs. Please refer to Table: \ref{Sub-Task-2-Test} for detailed results related to this validation.

\section{Conclusion}
For INLG 2022, we created a system to predict the Average Rating of synthetically generated Hinglish sentences (Sub-Task 1) \& Disagreement score for the same (Sub-Task 2). We didn't use any outside information. We have used GLOVE embedding for English and Hindi sentences. And for Hinglish sentences we have used multi label vectors.

\bibliography{custom}

\begin{thebibliography}{13}
\expandafter\ifx\csname natexlab\endcsname\relax\def\natexlab#1{#1}\fi

\bibitem[{Das and Gamb{\"a}ck(2014)}]{das2014identifying}
Amitava Das and Bj{\"o}rn Gamb{\"a}ck. 2014.
\newblock Identifying languages at the word level in code-mixed indian social
  media text.

\bibitem[{Li et~al.(2006)Li, Yaman, Lee, Ma, Tong, Zhu, and Li}]{4013499}
Jinyu Li, Sibel Yaman, Chin-hui Lee, Bin Ma, Rong Tong, Donglai Zhu, and
  Haizhou Li. 2006.
\newblock \href {https://doi.org/10.1109/ODYSSEY.2006.248082} {Language
  recognition based on score distribution feature vectors and discriminative
  classifier fusion}.
\newblock In \emph{2006 IEEE Odyssey - The Speaker and Language Recognition
  Workshop}, pages 1--5.

\bibitem[{Linhares~Pontes et~al.(2018)Linhares~Pontes, Huet, Linhares, and
  Torres-Moreno}]{unknown}
Elvys Linhares~Pontes, Stéphane Huet, Andréa Linhares, and Juan-Manuel
  Torres-Moreno. 2018.
\newblock Predicting the semantic textual similarity with siamese cnn and lstm.

\bibitem[{Liu et~al.(2021)Liu, Wei, and Vosoughi}]{liu2021language}
Ruibo Liu, Jason Wei, and Soroush Vosoughi. 2021.
\newblock Language model augmented relevance score.
\newblock \emph{arXiv preprint arXiv:2108.08485}.

\bibitem[{Merlo et~al.(2003)Merlo, Henderson, Schneider, and Wehrli}]{article}
Paola Merlo, James Henderson, Gerold Schneider, and Eric Wehrli. 2003.
\newblock \href {https://doi.org/10.13092/lo.17.788} {Learning document
  similarity using natural language processing}.
\newblock \emph{Linguistik online}, 17.

\bibitem[{Patra et~al.(2018)Patra, Das, and Das}]{patra2018sentiment}
Braja~Gopal Patra, Dipankar Das, and Amitava Das. 2018.
\newblock Sentiment analysis of code-mixed indian languages: An overview of
  sail\_code-mixed shared task@ icon-2017.
\newblock \emph{arXiv preprint arXiv:1803.06745}.

\bibitem[{Pratapa et~al.(2018)Pratapa, Choudhury, and
  Sitaram}]{pratapa2018word}
Adithya Pratapa, Monojit Choudhury, and Sunayana Sitaram. 2018.
\newblock Word embeddings for code-mixed language processing.
\newblock In \emph{Proceedings of the 2018 conference on empirical methods in
  natural language processing}, pages 3067--3072.

\bibitem[{Srivastava and Singh(2020)}]{srivastava2020phinc}
Vivek Srivastava and Mayank Singh. 2020.
\newblock Phinc: A parallel hinglish social media code-mixed corpus for machine
  translation.
\newblock \emph{arXiv preprint arXiv:2004.09447}.

\bibitem[{Srivastava and
  Singh(2021{\natexlab{a}})}]{srivastava-singh-2021-hinge}
Vivek Srivastava and Mayank Singh. 2021{\natexlab{a}}.
\newblock \href {https://doi.org/10.18653/v1/2021.eval4nlp-1.20} {{H}in{GE}: A
  dataset for generation and evaluation of code-mixed {H}inglish text}.
\newblock In \emph{Proceedings of the 2nd Workshop on Evaluation and Comparison
  of NLP Systems}, pages 200--208, Punta Cana, Dominican Republic. Association
  for Computational Linguistics.

\bibitem[{Srivastava and
  Singh(2021{\natexlab{b}})}]{srivastava-singh-2021-quality}
Vivek Srivastava and Mayank Singh. 2021{\natexlab{b}}.
\newblock \href {https://aclanthology.org/2021.inlg-1.34} {Quality evaluation
  of the low-resource synthetically generated code-mixed {H}inglish text}.
\newblock In \emph{Proceedings of the 14th International Conference on Natural
  Language Generation}, pages 314--319, Aberdeen, Scotland, UK. Association for
  Computational Linguistics.

\bibitem[{Srivastava and Singh(2022)}]{srivastava2022code}
Vivek Srivastava and Mayank Singh. 2022.
\newblock Code-mixed nlg: Resources, metrics, and challenges.
\newblock In \emph{5th Joint International Conference on Data Science \&
  Management of Data (9th ACM IKDD CODS and 27th COMAD)}, pages 328--332.

\bibitem[{Wang et~al.(2018)Wang, Liu, Sakaji, Ito, Izumi, Tsubouchi, and
  Yamashita}]{jrfm11010008}
Zhouhao Wang, Enda Liu, Hiroki Sakaji, Tomoki Ito, Kiyoshi Izumi, Kota
  Tsubouchi, and Tatsuo Yamashita. 2018.
\newblock \href {https://doi.org/10.3390/jrfm11010008} {Estimation of
  cross-lingual news similarities using text-mining methods}.
\newblock \emph{Journal of Risk and Financial Management}, 11(1).

\bibitem[{Yadav and Chakraborty(2020)}]{yadav2020unsupervised}
Siddharth Yadav and Tanmoy Chakraborty. 2020.
\newblock Unsupervised sentiment analysis for code-mixed data.
\newblock \emph{arXiv preprint arXiv:2001.11384}.

\end{thebibliography}
\bibliographystyle{acl_natbib}
\nocite{*}




\end{document}